%% file: arxiv.tex
\documentclass[10pt,twocolumn,letterpaper]{article}

\usepackage[final,algorithms]{wacv}
\usepackage{times}
\usepackage{epsfig}
\usepackage{graphicx}
\usepackage{amsmath}
\usepackage{amssymb}
\usepackage{multirow}
\usepackage{booktabs}
\usepackage{algorithm}
\usepackage{algpseudocode}
\usepackage{subcaption}
\usepackage{color,soul}
\usepackage[percent]{overpic}
\usepackage[accsupp]{axessibility}  

\input{helpers/math}

\usepackage[pagebackref,breaklinks,colorlinks]{hyperref}

\usepackage[capitalize]{cleveref}
\crefname{section}{Sec.}{Secs.}
\Crefname{section}{Section}{Sections}
\Crefname{table}{Table}{Tables}
\crefname{table}{Tab.}{Tabs.}

\newcommand{\methodabbr}{OSA-DAS}
\newcommand{\methodfullname}{Occlusion Sensitivity Analysis with Deep Feature Augmentation Subspace}
\newcommand{\modeloutput}{deep feature vector}

\begin{document}

\title{Occlusion Sensitivity Analysis with Augmentation Subspace Perturbation
in Deep Feature Space}

\author{Pedro H. V. Valois\\
University of Tsukuba\\
{\tt\small pedro@cvlab.cs.tsukuba.ac.jp}
\and
Koichiro Niinuma\\
Fujitsu Research of America\\
{\tt\small kniinuma@fujitsu.com}
\and
Kazuhiro Fukui\\
University of Tsukuba\\
{\tt\small kfukui@cs.tsukuba.ac.jp}
}

\date{}

\maketitle

\input{sections/abstract}


\section{Introduction} \label{sec:into}
\input{sections/introduction}

\section{Related Work} \label{sec:relatedwork}
\input{sections/related_work}


\section{Methods} \label{sec:methods}
\input{sections/methods}

\section{Experiments} \label{sec:experiments}
\input{sections/experiments}

\section{Conclusion} \label{sec:conclusion}
\input{sections/conclusions}

{\small
\bibliographystyle{ieee_fullname}
\bibliography{zotero}
}

\newpage

\section{Appendix} \label{sec:supplementary}
\input{supplementary/supplementary}

\end{document}

%% file: helpers/math.tex
\usepackage{amsmath}
\usepackage{amsthm}
\usepackage{amsfonts}
\usepackage{amssymb}
\usepackage{mathtools}
\usepackage{enumitem}
\usepackage{makecell}

\newcommand*\autoop{\left(}
\newcommand*\autocp{\right)}
\newcommand*\autoob{\left[}
\newcommand*\autocb{\right]}
\AtBeginDocument {%
   \mathcode`( 32768
   \mathcode`) 32768
   \mathcode`[ 32768
   \mathcode`] 32768
   \begingroup
       \lccode`\~`(
       \lowercase{%
   \endgroup
       \let~\autoop
   }\begingroup
       \lccode`\~`)
       \lowercase{%
   \endgroup
       \let~\autocp
   }\begingroup
       \lccode`\~`[
       \lowercase{%
   \endgroup
       \let~\autoob
   }\begingroup
       \lccode`\~`]
       \lowercase{%
   \endgroup
       \let~\autocb
}}
\makeatletter
\AtBeginDocument {%
          \def\resetMathstrut@{%
           \setbox\z@\hbox{\the\textfont\symoperators\char40}%
           \ht\Mathstrutbox@\ht\z@ \dp\Mathstrutbox@\dp\z@}%
}%
\makeatother

\newcolumntype{L}{>{$}l<{$}}

\usepackage{mathtools}

\DeclarePairedDelimiterX\braket[2]{\langle}{\rangle}{#1 \delimsize\vert #2}
\DeclarePairedDelimiterX\brakaket[3]{\langle}{\rangle}{#1 \delimsize\vert #2 \delimsize\vert #3}

\newcommand{\tbraket}[2]{$\braket{1}{2}$}

\theoremstyle{remark}

\theoremstyle{remark}
\newtheorem{remark}{Remark}

\theoremstyle{definition}
\newtheorem{definition}{Definition}[section]

\theoremstyle{definition}

\theoremstyle{definition}

\newtheorem{theorem}{Theorem}[section]

\newtheorem{lemma}{Lemma}[section] 
\newtheorem{proposition}{Proposition}

\newcommand{\indices}[3][\,]{
  {#2_1 #1 #2_2 #1 \dotsb #1 #2_{#3}}
}

%% file: sections/abstract.tex
\begin{abstract}

Deep Learning of neural networks has gained prominence in multiple life-critical applications like medical diagnoses and autonomous vehicle accident investigations. However, concerns about model transparency and biases persist. Explainable methods are viewed as the solution to address these challenges. In this study, we introduce the \methodfullname{} (\methodabbr{}), a novel perturbation-based interpretability approach for computer vision. While traditional perturbation methods make only use of occlusions to explain the model predictions, \methodabbr{} extends standard occlusion sensitivity analysis by enabling the integration with diverse image augmentations. Distinctly, our method utilizes the output vector of a DNN to build low-dimensional subspaces within the \modeloutput{} space, offering a more precise explanation of the model prediction. The structural similarity between these subspaces encompasses the influence of diverse augmentations and occlusions. We test extensively on the ImageNet-1k, and our class- and model-agnostic approach outperforms commonly used interpreters, setting it apart in the realm of explainable AI. \footnote{This work has been accepted for IEEE/CVF Winter Conference on Applications of Computer Vision (WACV) 2024}
\end{abstract}

%% file: sections/introduction.tex
Interpretability in deep learning provides insights into the complex operations of deep neural networks (DNNs), which often seem like ``black boxes'' due to their intricate structures. There's a growing demand for interpreters, tools that decode the influence of input features on a DNN's decisions, especially in critical areas like healthcare and autonomous vehicles. Effective explanations enhances user trust, highlight model biases and also its strengths, fostering wider acceptance of these systems~\cite{adebayo_sanity_2018,hooker_benchmark_2019,the_white_house_president_2023}.

Within this field, perturbation-based methods are those which attempt to explain the machine learning model by connecting input modifications with output changes to construct an explanation heatmap, \ie, a 2D attribution matrix indicating the responsibility of each input pixel to the model prediction~\cite{chockler_explanations_2021,halpern_causes_2005-1,halpern_causes_2005,halpern_modification_2015}. In that sense, occlusion is one of such methods, measuring the responsibility of each pixel by replacing image regions with a given baseline, \eg, setting it to zero, and measuring output variations~\cite{fleet_visualizing_2014, petsiuk_rise_2018}. Nevertheless, careless occlusion likely generates images which are outside of the training data's distribution, leading to unfair comparisons and fragile visualizations~\cite{hooker_benchmark_2019}.

In order to address this shortcoming, we propose a novel interpretability framework that integrates naïve occlusion with other common image augmentations employed during model training. Our proposal hinges on a simple premise: if data augmentations are pivotal in model training, they can be equally instrumental in enhancing interpretability as the model's reaction to augmentations is a viable path to understand its decision-making process. 
However, seamlessly integrating these augmentations is not trivial. For example, \textit{if jittering the color of an image changes the model output, how to pinpoint which region was most affected by it?} 

\begin{figure}[!t]
\begin{center}
\begin{overpic}[width=1.04\linewidth]{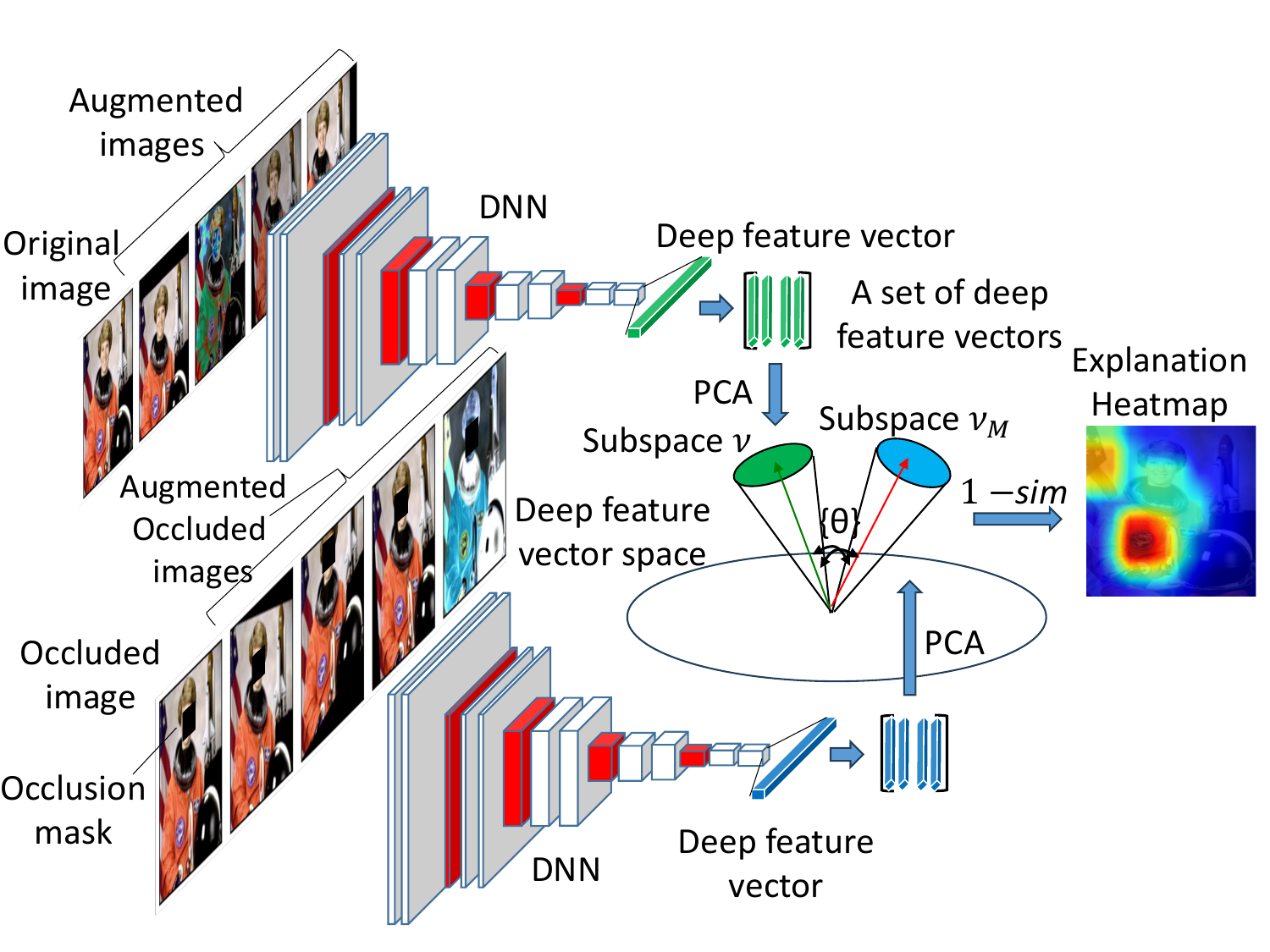}
 
\end{overpic}
\end{center}
   \caption{\methodabbr{} Overview: Subspace \(\mathcal{V}\) is derived from the augmented input image, while \(\mathcal{V_M}\) originates from its occluded counterpart. Both are derived from the principal component analysis (PCA) of a DNN's \modeloutput{}. The orthogonal degree~\cite{fukui_difference_2015,fukui_subspace_2020} between \(\mathcal{V}\) and \(\mathcal{V_M}\) quantifies the occlusion's effect and shapes the explanation heatmap. Multiple occlusion augmentation subspaces are used to capture diverse facets of the input's representation. Their combined relationships offer a holistic view of occlusion impacts, producing a detailed heatmap.}
\label{fig:rosa-overview}
\end{figure}

Thus, a challenge arises when trying to determine the specific impact of an augmentation. 
Our approach relies on the DNN \modeloutput{} from the final layer before the classification head. We feed both the original images and their augmented variants (with or without occlusion) to CNNs or Vision Transformers. This yields two sets of \modeloutput{}s: one from original/augmented images without occlusion and another from their occluded counterparts, as depicted in \cref{fig:rosa-overview}. 

We then compactly represent each set as a low-dimensional subspace in the deep feature vector space by applying Principal Component Analysis (PCA) without data centering to the set. Two subspaces ${\mathcal{V}}_M$ and $\mathcal{V}$ are generated from the sets extracted from images with and without occlusion, respectively. The core idea of our proposal is to measure the small perturbation due to the occlusion by the structural similarity $Sim$ between ${\mathcal{V}}_M$ and $\mathcal{V}$, which is defined using the multiple canonical angles $\{\theta\}$ between the subspaces \cite{fukui_difference_2015}. A larger subspace distance (orthogonal degree), $1 - Sim$, signifies that the occluded region is crucial for classification. This subspace representation method streamlines the process of merging multiple augmentation influences, offering a straightforward and robust metric of structural difference in the \modeloutput{} space.


Overall, our contributions are as follows:

\begin{enumerate}
    \item We introduce a novel interpretability framework able to leverage any data augmentation to improve DNNs prediction explanation, shown in \cref{fig:rosa-overview}.
    \item We leverage subspace representations~\cite{fukui_difference_2015} with the \modeloutput{} in explanation methods. This approach facilitates a more granular understanding of the model's behavior and offers a robust explanation.
    \item We optimize our algorithm by designing a better random masking routine, which proposes better occlusions, allowing for a faster convergence.
    \item We present a new interpretability metric named minimal size, which relies on causality theory~\cite{halpern_causes_2005} to measure how close the explanation heatmap is to the actual cause of the model prediction.
\end{enumerate}

%% file: sections/related_work.tex
The visualization of deep learning models decision-making process has become a vital research area, given the complex and often opaque nature of neural networks. Many methods have been introduced to shed light on how DNNs arrive at specific predictions. Gradient-based methods generate visualizations from the model output derivative with respect to the input image~\cite{simonyan_deep_2013,sundararajan_axiomatic_2017}. Activation-based methods~\cite{selvaraju_grad-cam_2016,montavon_explaining_2017,chattopadhay_grad-cam_2018,smilkov_smoothgrad_2017,muhammad_eigen-cam_2020} build upon gradients but take into consideration common properties of the network structure, which improves output. These techniques can compute heatmaps quite fast yet many times lack explainability, showing many similarities to an edge detector~\cite{adebayo_sanity_2018,kokhlikyan_investigating_2021}. 

Additionally, given the recent developments in transformers~\cite{vaswani_attention_2017,dosovitskiy_image_2021,liu_swin_2021,liu_swin_2022}, a new family of attention-based interpreters has been proposed~\cite{abnar_quantifying_2020,chefer_transformer_2021}, in which the attention weights from multiple layers are used to compute explanations. These methods demonstrate elevated interpretability capacity, but they are architecture-specific.

On the other hand, perturbation-based methods make minimal assumptions about the nature of the model itself and exactly for that reason show increased ability in explaining any kind of machine learning model. The basic perturbation method, Occlusion Sensitivity Analysis (OSA)~\cite{fleet_visualizing_2014}, is actually quite straightforward. First, it measures the slight variation of the class score to occlusion in different regions of an input image using small perturbations of the image. Then, the resultant variation of each region is summarized as a heatmap of the input image. Other methods propose extensions to this idea by introducing new ways to generate the optimal occlusions~\cite{petsiuk_rise_2018,fong_interpretable_2017,fong_understanding_2019,uchiyama_visually_2023} or on how to compute their contributions~\cite{chockler_explanations_2021}.

Nevertheless, these methods are unable to explain the whole range of possibilities that can lead to a prediction, and have been criticized for analyzing the model on a different data distribution on which it was trained~\cite{hooker_benchmark_2019}. In that sense, the robustness of visual explanations to common data augmentation techniques, such as occlusions, has been studied. \cite{tetkova_robustness_2023} analyzed the response of post-hoc visual explanations to natural data transformations. They found significant differences in robustness depending on the type of transformation, with some techniques demonstrating more stability. Similarly, \cite{won_analyzing_2023} explored the relationship between data augmentation strategies and model interpretability, revealing that models trained with mixed sample data augmentation showed lower interpretability, particularly with CutMix~\cite{yun_cutmix_2019} and SaliencyMix~\cite{uddin_saliencymix_2020} augmentations. Moreover, \cite{arthur_oliveira_santos_impact_2022} proposes an augmentation method leveraging multiple interpreters, thereby enhancing model robustness against noise or occlusions.
This highlights the complex relationship between augmentation techniques and interpretability, raising caution for their adoption in critical applications. However, it's noteworthy that while these works analyze the impact of augmentations on explanations, as far as we know, none proposes an interpreter that leverages augmentation specifically to improve explanation trustworthiness.

%% file: sections/methods.tex
In this section, we introduce our original method and metric. Details can be found in the supplementary material.

\subsection{Occlusion with Augmentation Subspaces}

Traditional occlusion sensitivity analysis (OSA) computes explanation heatmaps by replacing image regions with a given baseline (masking it to 0), and measuring the score difference in the output~\cite{fleet_visualizing_2014, petsiuk_rise_2018}. While this technique is cost-effective, occluded images originate from a distinct distribution from the one which the model was trained on. Thus, discerning whether the performance dip arises from this distributional shift or due to the responsibility of the occluded regions becomes ambiguous.

On the other hand, data augmentation (including random occlusions) have been used in most state-of-the-art models during training~\cite{balestriero_data-augmentation_2022,cubuk_randaugment_2020,muller_trivialaugment_2021}. Therefore, we expect a more accurate interpretation could be performed if the model uses augmentations closer to the real training distribution.

With that in mind, we devise a technique that adapts OSA to using any data augmentation routine in an independent way by leveraging subspaces of \modeloutput{}s. 

\subsubsection{Data Augmentation Methods}

\methodfullname{} (\methodabbr{}) utilizes data augmentation methods to foster more distinctive \modeloutput{}s that can be leveraged for enhanced interpretability.

In the realm of data augmentation, there exist prominent state-of-the-art routines that have revolutionized the process. For instance, RandAugment~\cite{cubuk_randaugment_2020} is an automated data augmentation approach that streamlines the selection of transformations through two hyperparameters: $n_{ops}$, denoting the number of sequential augmentation transformations, and $mag$, representing the magnitude of these transformations. The transformations span from simple affine transformations, such as rotation and translation to more intricate operations such as color jittering and auto contrast.

On the other hand, TrivialAugment~\cite{muller_trivialaugment_2021} presents an elegant yet powerful approach to automatic augmentation. It stands out due to its simplicity, requiring no parameters and applying a singular augmentation to each image. Despite its minimalist design, it has demonstrated its prowess, outperforming more complex augmentation techniques.

Central to our method is its adaptability and versatility. We chose the aforementioned augmentations in our experiments given they represent the pinnacle of current techniques, but our proposed framework is inherently flexible. It is designed to seamlessly integrate with any data augmentation routine, be it RandAugment~\cite{cubuk_randaugment_2020}, TrivialAugment~\cite{muller_trivialaugment_2021} or else that best fits the explanation goal of the task at hand.

\subsubsection{Deep Feature Augmentation Subspace}

The addition of any data augmentation to perturbation-based interpretability is not trivial, and we opt to use sets of augmented inputs around each occlusion.

Consider that an image $\mathbf{x}$ is fed into the model \(f()\). In this paper, the output $\mathbf{v} = f(\mathbf{x})$ is referred to as a \modeloutput{} in a $k$-dimensional vector space. For each occlusion $\mathbf{M}$, we generate a set of deep feature vectors corresponding to augmented images with occlusions, and then represent compactly the set by a subspace $\mathcal{V}_\mathbf{M} \subset {\mathbb{R}}^k$ for the specific occlusion. The same is performed for the original input image, which builds the reference subspace $\mathcal{V} \subset {\mathbb{R}}^k$.

The orthonormal basis, $\mathbf{V}$ and $\mathbf{V}_\mathbf{M} \in {\mathbb{R}}^{k{\times}d}$, of the $d$-dimensional subspaces \( \mathcal{V} \) and \( \mathcal{V}_\mathbf{M} \) are calculated by applying Principal Component Analysis (PCA) without data centering to each set of deep feature vectors. More concretely, they can be obtained as the eigenvectors corresponding to several largest eigenvalues of auto-correlation matrix $\sum_{i=1}^{m} \mathbf{v}_i \mathbf{v}_i^T \in {\mathbb{R}}^{k{\times}k}$, where $m$ is the number of applied augmentation types.

\subsubsection{Structural Similarity between Two Subspaces}

The relationship between two $d$-dimensional subspaces in ${\mathbb{R}}^k$ is defined by a set of $d$ canonical angles $\{\theta_i\}_{i=1}^{d}$ between them. 
They can be obtained by applying singular value decomposition (SVD) to $\mathbf{V}^T \mathbf{V}_\mathbf{M}$, where $\mathbf{V}$ and $\mathbf{V}_\mathbf{M} \in {\mathbb{R}}^{k{\times}d}$ are the orthonormal basis \cite{fukui_difference_2015}. The $\cos\theta_i$ of the $i$-th smallest canonical angle $\theta_i$ is the $i$-th largest singular value:
\begin{equation}
    \cos\theta_i = \sigma_{i}(\mathbf{V}^T \mathbf{V}_\mathbf{M}) ,
\end{equation}
where $\sigma_{i}(\cdot)$ returns the matrix $i$-th largest singular value.

The structural similarity between two subspaces is defined as the sum of the square of the cosines of the first $n_c$ canonical angles, where $n_c$ is a hyperparameter indicating how much information from each subspace is to be considered~\cite{fukui_difference_2015, fukui_subspace_2020}. However, in our method, we need a measurement of subspace distance, which can be used as a proxy for the degree of responsibility $r$ of each occlusion. Thus, we introduce the subspace distance, i.e., orthogonal degree, \cite{fukui_difference_2015} defined by the following equation:

\begin{equation}
    r({\mathbf{M}}) = 1 - \sum_i^{n_c} (\sigma_{i}(\mathbf{V}^T \mathbf{V}_\mathbf{M}))^2 .
\end{equation}

\subsubsection{Speedup by improved masking} \label{subsubsec:anchors}

\begin{figure}[!tb]
\begin{center}
\vspace{4mm}
\begin{overpic}[width=0.99\linewidth]{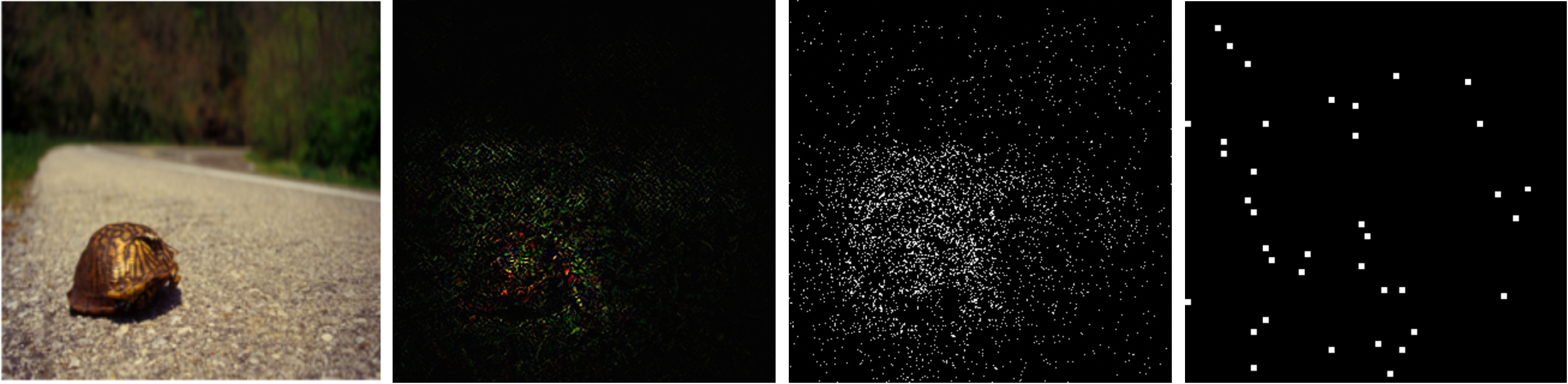}
 
 \put (7,-3) {\small Image}

 \put (22,29) {\small $\displaystyle\nabla f$}
 \put (22,25) {\large $\curvearrowright$}

 \put (31,-3) {\small Gradient}

 \put (42,29) {\small Sampling}
 \put (47,25) {\large $\curvearrowright$}

 \put (55,-3){\small Samples}

 \put (69,29) {\small Filtering}
 \put (73,25) {\large $\curvearrowright$}
 
 \put (81,-3){\small Anchors}
\end{overpic}
 
\vspace{-4mm}
\end{center}
   \caption{Mask anchor point selection via gradient sampling. The image gradient is produced on inference time, which is then used to sample anchor points. Anchor points too close to each other are filtered out. (anchors size is increased for visibility)}
\label{fig:rosa-anchors}
\end{figure}
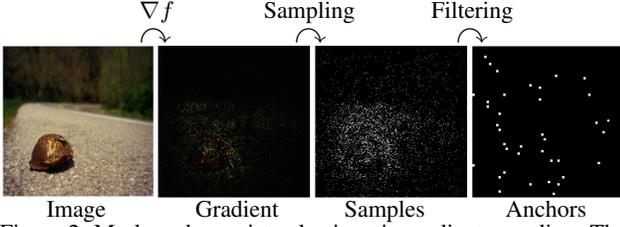

\methodabbr{} enhances OSA by incorporating more information, albeit at a higher computational cost. Essentially, perturbation-based interpretability is akin to a Monte Carlo approach for estimating machine learning models. The efficiency of this method can be improved by proposing better masks, thereby reducing the number of required masks, as seen in \cite{uchiyama_visually_2023}. One straightforward strategy to devise superior masks is to utilize the model's gradient concerning the input image as weights. Albeit the simplicity, gradients are know to be noisy and not always indicate the most relevant features~\cite{sundararajan_axiomatic_2017,smilkov_smoothgrad_2017}, yet can be leveraged to sample the mask anchor points using a multinomial distribution. However, direct sampling often results in highly overlapping masks. To address this, we filter out those with substantial overlapping mask areas, as illustrated in \cref{fig:rosa-anchors}.

\subsubsection{Algorithmic generation of Explanation heatmaps}

\begin{algorithm}
\caption{\methodfullname{} (\methodabbr{})} \label{alg:rosa-sm}
\begin{algorithmic}
\Require $\mathbf{x}\gets\text{image}, f{\gets}\text{model}$, $\tau_i{\gets} i\text{-th augmentation}$
\State $n_m \gets \text{number of masks}$
\State $n_a \gets \text{number of augmentations}$
\State $n_c \gets \text{number of canonical angles}$
\State $l \gets \text{mask size}$

\State $\mathcal{V} \gets \{ \}$
\For{$i \gets 1$ to $n_a$}
    \State ${\mathbf{x}}_t \gets \tau_i(\mathbf{x})$ 
    \State Insert the normalized $f({\mathbf{x}}_t) \in {\mathbb{R}}^{k}$ in $\mathcal{V}$
     \EndFor
    \State $\mathbf{V} \gets PCA(\mathcal{V})$

\State $\mathbf{H} \gets 0$
\For{$i \gets 1$ to $n_m$}
    \State $\mathbf{M} \gets \text{mask}(i, \mathbf{x}.shape, l)$ 
    \State $\mathcal{V}_\mathbf{M} \gets \{ \}$
    \State ${\mathbf{x}}^\mathbf{M} \gets \mathbf{x} \odot \mathbf{M}$
    \For{$j \gets 1$ to $n_a$}
        \State ${\mathbf{x}}^\mathbf{M}_t \gets {\tau}_j({\mathbf{x}}^\mathbf{M})$
        \State Insert the normalized $f({\mathbf{x}}^\mathbf{M}_t)\in {\mathbb{R}}^{k}$ in $\mathcal{V}_\mathbf{M}$ 
    \EndFor

    \State $\mathbf{V}_\mathbf{M} \gets PCA(\mathcal{V}_\mathbf{M})$
    \State $r \gets 1 - \sum_k^{n_c}(\sigma_k(\mathbf{V}^T\mathbf{V}_\mathbf{M}))^2$
    \State $\mathbf{H} \gets \mathbf{H} + (1 - \mathbf{M}) r$ 
\EndFor
\end{algorithmic}
\Return $\frac{\mathbf{H}}{\sum \mathbf{H}}$ 
\end{algorithm}

The presented ideas for the basis of our method is fully presented in \Cref{alg:rosa-sm}. Our OSA-DAS begins by sampling a set of augmentations on the original image. It then constructs a subspace, \(\mathcal{V}\), which captures the model outputs for these augmented images. For each occlusion applied to the image, a similar subspace, \(\mathcal{V}_\mathbf{M}\), is formed. The goal is then to compare the two subspaces, \(\mathcal{V}\) and \(\mathcal{V}_\mathbf{M}\), to understand the significance of the occluded region.

\begin{enumerate}
    \item \textbf{Initialization:} Let $f$ be a deep learning model that outputs a $k$-dimensional deep feature vector extracted from an input image. Let $\mathbf{x}$ be an input image and $\{{\tau}_i(\mathbf{x})\}_{i=1}^{n_a}$ a set of its augmentations. Besides, set parameters for the number of masks $n_m$, augmentations $n_a$, number of canonical angles $n_c$, and mask size \(l\).  
    \item \textbf{Construct the Reference Subspace \(\mathcal{V}\):}
    For $i$-th augmentation \(\tau_i\):
    \begin{enumerate}
        \item Feed augmented image $\tau_i(\mathbf{x})$ into the model \(f\).
        \item Normalize the length and store the deep feature vector $f(\tau_i(\mathbf{x})) \in {\mathbb{R}}^k $ in an array.
    \end{enumerate}
    We conduct the above process over all the augmentations, and then compute the orthonormal basis \(\mathbf{V} \in {\mathbb{R}}^{k{\times}d}\) of the \(\mathcal{V}\) subspace from the set of \modeloutput{}s ${\{f(\tau_i(\mathbf{x}))\}}_{i=1}^{n_a}$.
    
    \item \textbf{Sample Masks and Construct Occluded Subspaces:} For each mask generated:
    \begin{enumerate}
        \item Create occlusions in the image using the mask.
        \item For each occlusion, compute a basis \(\mathbf{V}_\mathbf{M} \in {\mathbb{R}}^{k{\times}d}\) of subspace \(\mathcal{V}_\mathbf{M}\) from the set of the $k$-dimensional feature vectors, ${\{f(\tau_i({\mathbf{x}}^M))\}}_{i=1}^{n_a}$, following the process in Step $2$.
    \end{enumerate}
    
    \item \textbf{Compute the orthogonal degree:} Measure the orthogonal degree $r(\mathbf{M})$ between the $d$-dimensional reference subspace \(\mathcal{V}\) and  occluded subspace \(\mathcal{V}_\mathbf{M}\).
    
    \item \textbf{Generate Explanation Heatmap:} Assign \(r({\mathbf{M}})\) to represent the significance of the occluded region $\mathbf{M}$. Combine these values across all occlusions ${\{{\mathbf{M}}_i\}}_{i=1}^{n_m}$ to form the heatmap $\mathbf{H}$. Normalize the heatmap to ensure values between $0$ and $1$.
\end{enumerate}

\begin{figure}[!tb]
    \centering
    \vspace{-3mm}
    \includegraphics[width=1.0\linewidth]{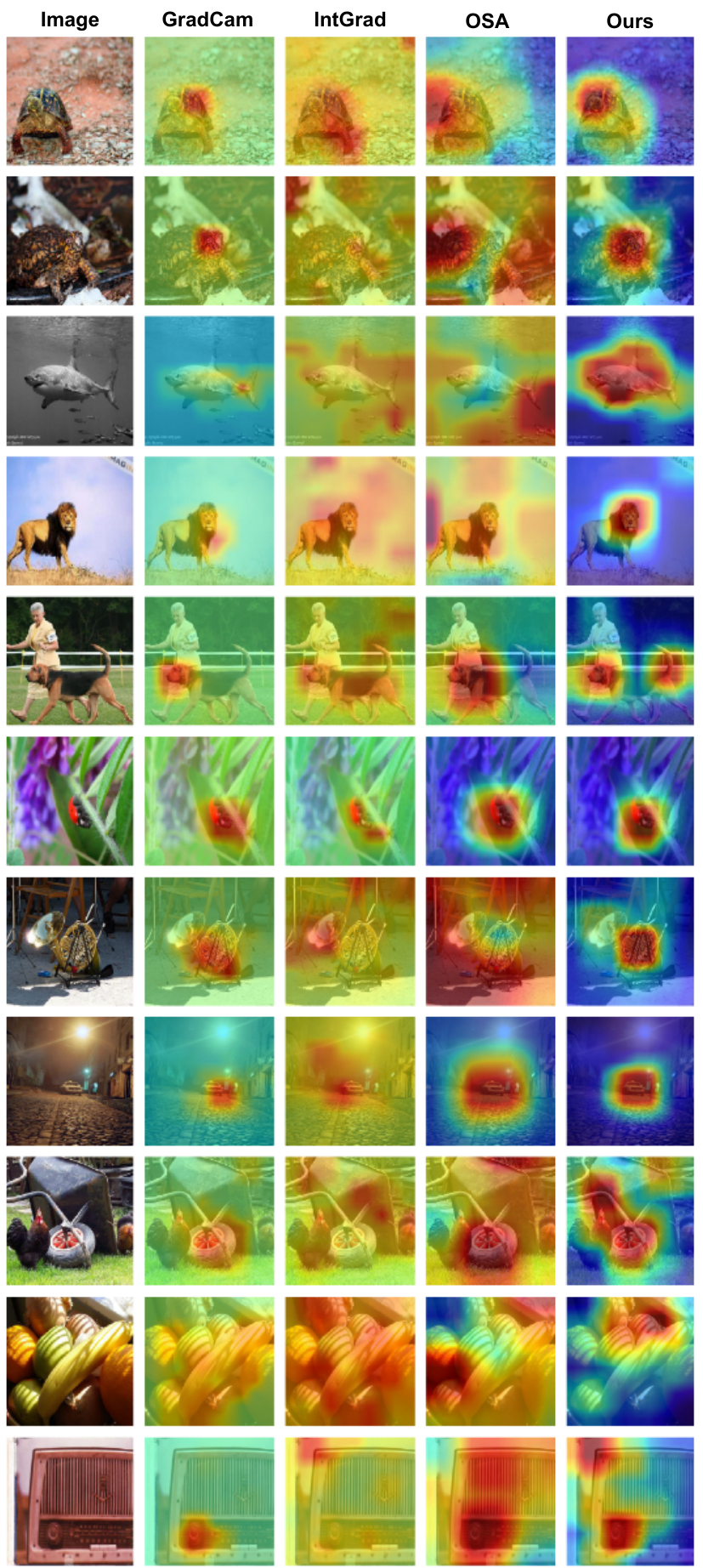}
    \caption{Explanation heatmaps visualizations for ResNet-50. Regions in red indicate the prediction causes. The proposed method generate concise and smooth explanation heatmaps, more in line to the general features the model is attending than other techniques.}
    \label{fig:heatmaps-resnet50}
    \vspace{-4mm}
\end{figure}

\subsection{Explanation and Metrics}

Even though the interpretability goal is to build clear visualizations of the machine learning model decision-making process, the comparison of interpreters at scale requires the application of metrics that can accurately measure the quality of the explanations~\cite{adebayo_local_2018}.

\subsubsection{Explanations} \label{subsubsec:explanations}
Given an input image $\mathbf{x}$, $\mathbf{S} = \mathbf{x} \odot \mathbf{M}$ indicates a masked subset of the input, where $\mathbf{M}$ is a binary mask and $\odot$ is the Hadamard product. Then, the explanation $\mathbf{E}$ is the minimal subset which has the same output as the original input.

\begin{equation} \label{eq:explanation}
    \mathbf{E}(f|\mathbf{x}) = \min_{|\mathbf{S}|} \mathbf{S} : f(\mathbf{S}) = f(\mathbf{x}), \text{ with } |\mathbf{S}| > 0 ,
\end{equation}
where $|\cdot|$ counts the number of unmasked pixels.

\cref{eq:explanation} is a general definition, and the nature of the model's output can vary depending on the algorithm. 
In this work, we want to build a class-agnostic method using \modeloutput{}s $\in \mathbb{R}^k$, which are extracted from the final layer before the classification head.

However, to compute the precise explanation using only \cref{eq:explanation} would require testing all possible subsets of pixels to ensure we have the minimal one~\cite{chockler_explanations_2021}. In that sense, real interpreters provide an approximate explanation heatmap $\mathbf{\tilde{E}}(f|\mathbf{x})$. This map is usually taken to be a description on how the model's predictions are influenced by each pixel~\cite{sundararajan_axiomatic_2017,halpern_causes_2005,chockler_explanations_2021}. In this work, we interpret these explanation heatmaps as probability distributions: they indicate the probability of each pixel in $\mathbf{x}$ belonging to the ideal explanation $\mathbf{E}(f|\mathbf{x})$. See the supplementary material for details.

\subsubsection{Evaluation Metrics} \label{subsubsec:metrics}

Many metrics have been proposed in interpretability literature, each offering different perspectives. In this paper, we chose to use multiple metrics to provide a more comprehensive measurement of the interpreter effectiveness. Deletion and insertion metrics~\cite{petsiuk_rise_2018} gauge the faithfulness of an explanation heatmap in representing a model's inferences. 

First, the deletion metric measures how rapidly the model's prediction probability decreases when pixels are deleted according to their heatmap significance. 

Conversely, the insertion metric evaluates how quickly the model's prediction probability escalates when pixels are inserted based on their heatmap significance. The performance of these metrics is quantified using the area under the curve ($AUC$), with the horizontal axis indicating the percentage of pixels deleted or inserted and the vertical axis representing the output probability of the model~\cite{chefer_transformer_2021}.

Although useful, we argue these metrics do not fundamentally align with the causality definition of explanation as per \cref{eq:explanation}. Also, their numbers are not so intuitive and most often than not it is difficult to link their values to any visible property of the heatmap.

\subsubsection{Minimal Size Metric} \label{subsubsec:minsize}

\cref{subsubsec:explanations} defines an explanation as the smallest set of pixels that still results in the same model output. 
Now, given an explanation heatmap $\mathbf{\Tilde{E}}(f|\mathbf{x})$, we can try to generate an explanation from it. If the heatmap is correct, the explanation must use the minimal number of pixels, and we can use this number as a viable metric of the proximity between the explanation heatmap and the model's output cause.

\begin{equation} \label{eq:minimalsize}
    s_{min}(\mathbf{\Tilde{E}}(f|\mathbf{x})) = \frac{|\mathbf{S}|}{|\mathbf{x}|}, \text{ with } f(\mathbf{S}) \approx f(\mathbf{x}),
\end{equation}

where $f(\mathbf{S}) \approx f(\mathbf{x})$ replaces the ideal equality $f(\mathbf{S}) = f(\mathbf{x})$ in \cref{eq:explanation} to make the metric less rigid while also improving numerical stability.

We stress that our metric is class-agnostic, which allows us to directly use \modeloutput{}s $f(\mathbf{S}) \in \mathbb{R}^k$ and $f(\mathbf{x}) \in \mathbb{R}^k$, while the deletion and insertion metrics are exclusively based on the change of the scalar class probability change measured with AUC. 


In practical terms, to compute this number we start from an empty set $\mathbf{S}$ and sequentially add pixels by order of importance, where pixel importance comes from $\mathbf{\Tilde{E}}$. During this process, we must reach a point such that $||f(\mathbf{S}) - f(\mathbf{x})||_1 \leq \delta$, where $||\cdot||_1 \leq \delta$ is an element-wise comparison within a fixed tolerance $\delta$. Then, the algorithm stops and the ratio $\frac{|\mathbf{S}|}{|\mathbf{x}|}$ is returned.

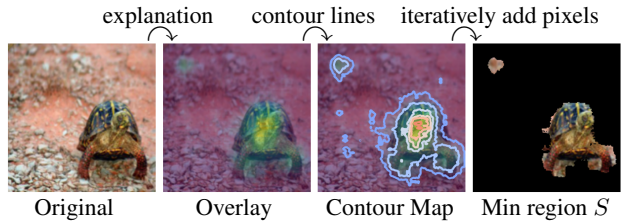
\begin{figure}[bt]
\begin{center}
\vspace{3mm}
\begin{overpic}[width=0.99\linewidth]{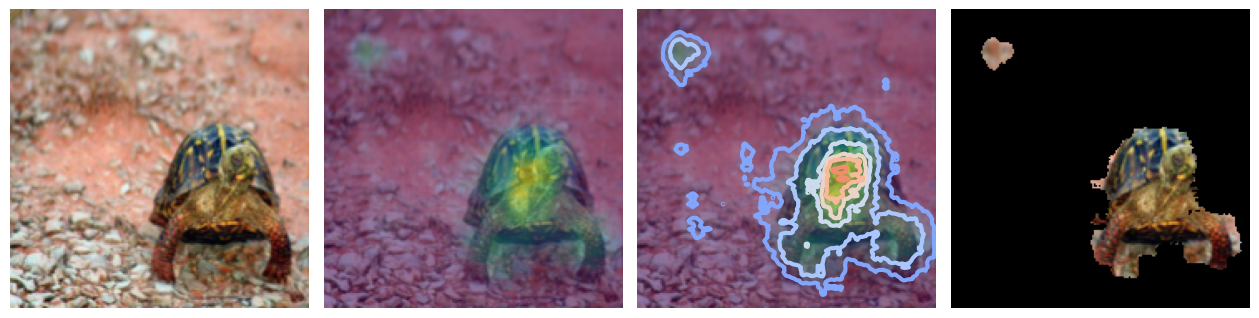}
 
 \put (5,-3) {\small Original}

 \put (16,28) {\small explanation}
 \put (23,25) {\large $\curvearrowright$}

 \put (31,-3) {\small Overlay}

 \put (40,28) {\small contour lines}
 \put (48,25) {\large $\curvearrowright$}

 \put (52,-3){\small Contour Map}

 \put (64,28) {\small iteratively add pixels}
 \put (72,25) {\large $\curvearrowright$}
 
 \put (77,-3){\small Min region $S$}
\end{overpic}
\vspace{-3mm}
\end{center}
   \caption{Simplified schema for computing the minimal size on a image-heatmap pair of $224 \times 224$ pixels with tolerance $\delta = 10^{-2}$.
   This simplified version decreases the iterations as follows: First, we divide the explanation heatmap into regions with the same importance level according to a contour map. Then, we introduce pixels from each region to the partial image in descending order of importance. We stop when the model's output of this partial image becomes very close to the one of the original image. The fraction of filled pixels in the partial image is the minimal size metric.}
\label{fig:minimal-size-process}
\vspace{-5mm}
\end{figure}

Although this works, the number of steps can be reduced by adding batches of pixels instead of one pixel at a time, as exemplified at \cref{fig:minimal-size-process}. Each batch is given from the contour map of $\mathbf{\Tilde{E}}$, which splits the heatmap into regions by the intensity of each pixel, and determines the number of pixels to be added at each step. 


Beyond that, notice a good interpreter metric should focus on evaluating only the explanation quality independently of model performance. This metric assesses the explanation's precision without being swayed by the model's accuracy while also providing a number that directly reflects the visual characteristics of the explanation. It's a clear and effective way to compare different interpreters' quality. See the supplementary material for more information.

\subsubsection{Overall performance metric} \label{subsubsec:overall}

While the Minimal Size metric offers a fresh perspective, it is essential to view it in conjunction with the currently used metrics for a holistic understanding. A more pivotal metric should thus be defined by balancing deletion, insertion and minimal size. We propose an Overall performance metric, building upon the work of \cite{zhang_group-cam_2021}, which combines insertion, deletion, and minimal size for a comprehensive evaluation.

\begin{equation} \label{eq:overall}
    overall = \dfrac{insertion - deletion}{minimal~size}
\end{equation}


\Cref{eq:overall} offers a more thorough understanding of interpreter performance. The incorporation of size in the denominator ensures a dimensionless metric, where both the numerator and denominator represent areas. This combined metric offers a balanced and insightful evaluation of the general interpreter performance, making it a more sensible evaluator of the general interpreter performance.

%% file: sections/experiments.tex
In this section, we present a comprehensive evaluation of our proposed methods through comparison with the conventional explanation methods. This includes qualitative comparisons of explanation heatmaps and a quantitative evaluation using deletion, insertion~\cite{petsiuk_rise_2018} and minimal size metrics.

\begin{figure}[!ht]
    \centering
    \includegraphics[width=1.00\linewidth]{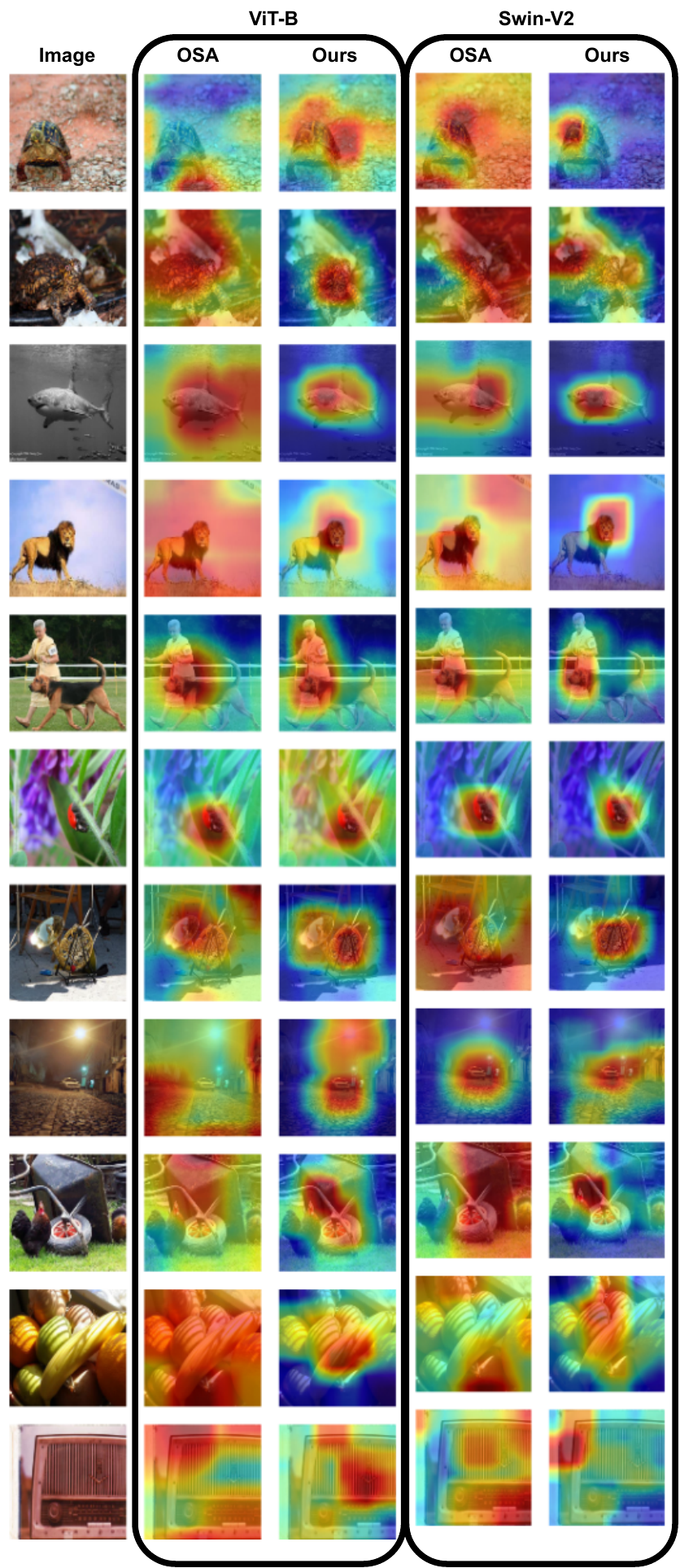}
    \caption{Explanation heatmaps visualizations for ViT-B and Swin-V2. Regions in red indicate the prediction causes.}
    \label{fig:heatmaps-transf}
    \vspace{-4mm}
\end{figure}

\subsection{Experiment Settings}

We employ ResNet-50~\cite{he_deep_2016}, ViT-B-14~\cite{dosovitskiy_image_2021} and Swin-V2~\cite{liu_swin_2021,liu_swin_2022} as classification models and assess the results on the validation set of ImageNet~\cite{russakovsky_imagenet_2015}, which comprises 50K images from 1000 classes and is used in explainable AI literature for evaluation~\cite{petsiuk_rise_2018,chockler_explanations_2021,chefer_transformer_2021}. The images are resized to $256 \times 256$ pixels and center cropped to $224 \times 224$ pixels.

For our method, we use masks of $64 \times 64$(=$l$) pixels in the image as described in \cref{subsubsec:anchors}, TrivialAugment~\cite{muller_trivialaugment_2021} as the augmentation routine, with 32(=$n_a$) augmentations per occlusion. The dimensions of the \modeloutput{} was 786 for all models. For comparison, we perform the evaluations together with Guided Grad-CAM~\cite{selvaraju_grad-cam_2016}, Integrated Gradients~\cite{sundararajan_axiomatic_2017} and OSA~\cite{fleet_visualizing_2014}, which are frequently employed interpreters of each major family of methods. Given our emphasis on developing model-agnostic methods, we refrained from comparing with non model-agnostic methods, such as \cite{abnar_quantifying_2020}, \cite{chefer_transformer_2021}, or expensive techniques like \cite{lundberg_unified_2017}.

We used the implementation of these methods provided by the Captum tool~\cite{kokhlikyan_captum_2020}. The batch of all experiments performed in this work, including ablations, took approximately one week to run on 8 V100 16Gb GPUS.

\begin{table*}[!ht]
    \centering
    \caption{Average Metric scores on ImageNet between ResNet-50, ViT-B and Swin-V2 models. For deletion and minimal size, lower is better ($\downarrow$). For insertion and overall, higher is better ($\uparrow$). \textbf{Bold} represents the best metric, while \underline{underline} is the second best. Occlusion and Ours have the same mask size, but the former uses a sliding window, so it generates much more masks.}
    \begin{tabular}{lcccc}
    \toprule
    Method & \textbf{Minimal Size ($\downarrow$)} & \textbf{Deletion ($\downarrow$)} & \textbf{Insertion ($\uparrow$)} & \textbf{Overall ($\uparrow$)} \\
    \midrule
    Guided Grad-CAM~\cite{selvaraju_grad-cam_2016} &  0.515 & \underline{0.298} & 0.289 & -0.034 \\
    Integrated Gradients~\cite{sundararajan_axiomatic_2017} & 0.518 & \textbf{0.234} & 0.267 & 0.123 \\
    Occlusion~\cite{fleet_visualizing_2014} & \underline{0.251} & 0.328 & \textbf{0.549} & \underline{0.880} \\
    \methodabbr{} (Ours) & \textbf{0.231} & 0.331 & \underline{0.539} & \textbf{0.901} \\
    \bottomrule
    \end{tabular}
    \label{tab:metric-comparisons-resumed}
\end{table*}


\subsection{Qualitative Results} \label{subsec:quali}

The visualization results of the explanation heatmaps are showcased in \cref{fig:heatmaps-resnet50} and \cref{fig:heatmaps-transf}. For the class-specific methods, we show the heatmap with respect to the predicted class. These images illustrate how other interpreters tend to generate noisy heatmaps, especially notable for traditional OSA on model misclassifications, which attributes inverted responsibility compared to \methodabbr{}.

Overall, the proposed method precisely captures the general features which the model attends to in a more stable manner, which facilitates model understanding and debugging. This resilience is likely due to its class-agnostic nature combined with the variety of feature comparisons enabled by the augmentation subspaces. These results suggest that \methodabbr{} is capable of selecting the most impactful regions for the model, regardless of mispredictions.

On the flip side, the increased memory cost restricts the maximum number of masks and augmentations that can be applied, posing a trade-off for achieving more robust and accurate explanations.

\subsection{Quantitative Results} \label{subsec:quanti}

Whereas \cref{subsec:quali} implies superiority of our proposed method, caution must be taken against sole reliance on visual assessments \cite{adebayo_sanity_2018}. The average evaluation results on the whole validation set of ImageNet are presented at \cref{tab:metric-comparisons-resumed} using the metrics of insertion, deletion (\cref{subsubsec:metrics}), minimal size (\cref{subsubsec:minsize}) and overall (\cref{subsubsec:overall}). Constant tolerance value $\delta = 10^{-2}$ is used. In deletion, the heatmap that accurately captures important individual pixels is highly valued, while for insertion, a heatmap presenting cohesive regions of importance is better evaluated~\cite{petsiuk_rise_2018,uchiyama_visually_2023}. Minimal size metric measures proximity of the explanation to the actual cause. Overall balances insertion, deletion and minimal size areas evenly. This experiment uses 32 iterations for insertion, deletion and minimal size. 

 Integrated Gradients~\cite{sundararajan_axiomatic_2017} and Grad-CAM~\cite{selvaraju_grad-cam_2016} focus too much on important pixels, but not on important regions, which optimizes deletion in detriment of other metrics. Occlusion shows excellent insertion performance because it exclusively focuses on the regions which impact the predicted class. On the other hand, our method showcases best overall performance, showing good results among all metrics. We argue this demonstrates it can explain the actual prediction cause in a more holistic and class-agnostic way than others. Further details on the performance for each model is shown in the supplementary material.

 In this context, it's noteworthy that class-specific methods, which consider specific priors for each class, are anticipated to perform better in insertion and deletion metrics compared to class-agnostic ones. This is because these methods priors (prediction probability) align with the same priors used in the evaluation metrics \cite{chefer_transformer_2021}. Regardless of such, our method still showcases comparable performance to OSA in spite of not being able to leverage such priors.

\subsection{Ablation}

We conducted an ablation study on our method's three key components: augmentations, masking, and subspace representations. These tests used 2500 ImageNet training images to ensure cost-efficiency and to maintain independence from analyses in \cref{subsec:quali} and \cref{subsec:quanti}. Tests used the ResNet-50 model, evaluating hyperparameters on a log2 scale until resource limits. We reported changes in the overall metric, defined in \cref{subsubsec:overall}. Other metrics showed consistent behaviors and led to the same conclusions.

\begin{figure}[!ht]
    \centering
    
    \begin{subfigure}[b]{0.49\columnwidth}
        \centering
        \begin{overpic}[width=\linewidth]{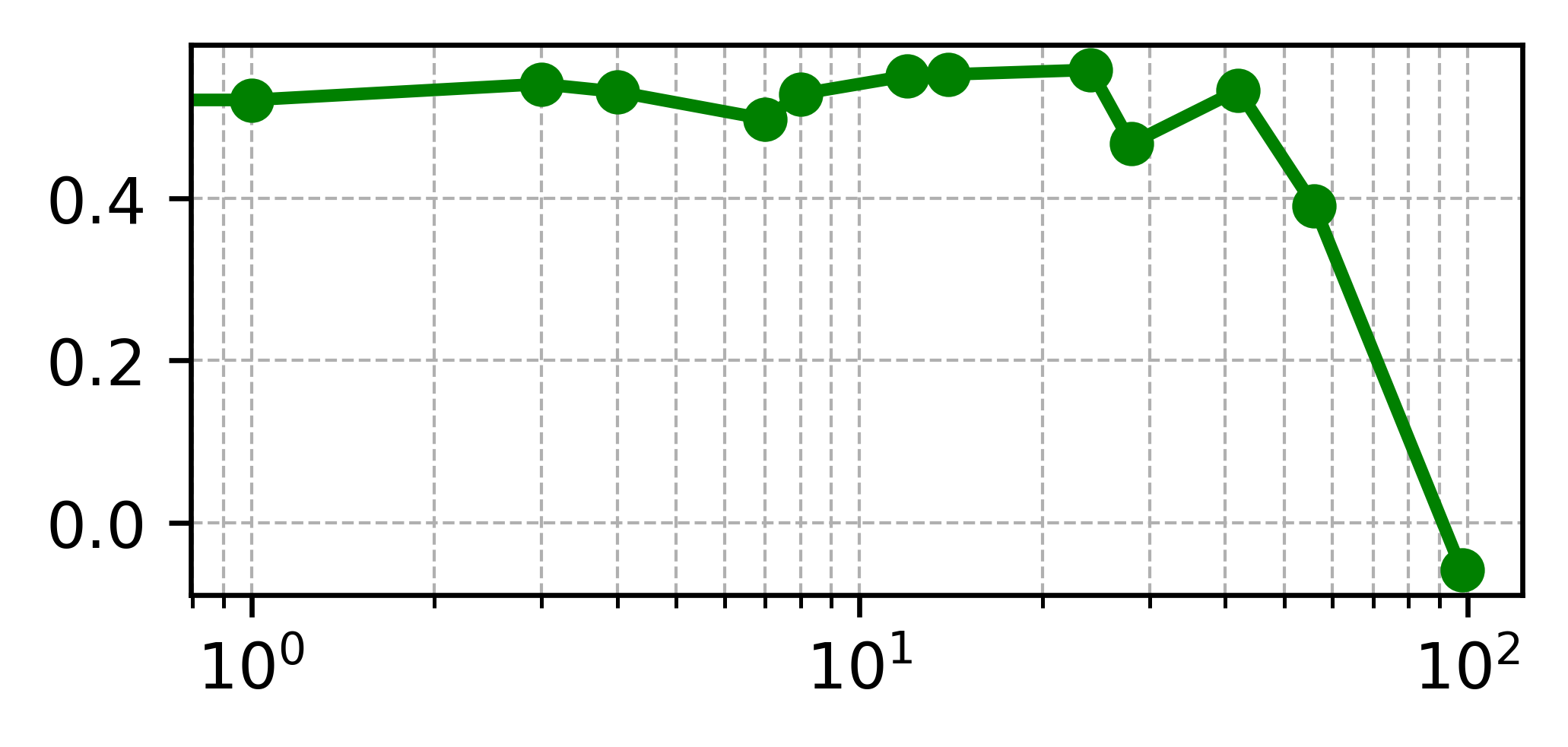}
         \put(-4,25){\makebox(0,0){\rotatebox{90}{Overall}}}
         \put (13,-4){Augmentation Strength}
        \end{overpic}
        \vspace{1mm}
        \caption{Augmentation strength influence over overall metric}
        \label{fig:ablation-aug-strength}
    \end{subfigure}
    \hfill
    \begin{subfigure}[b]{0.49\columnwidth}
        \centering
        \begin{overpic}[width=\linewidth]{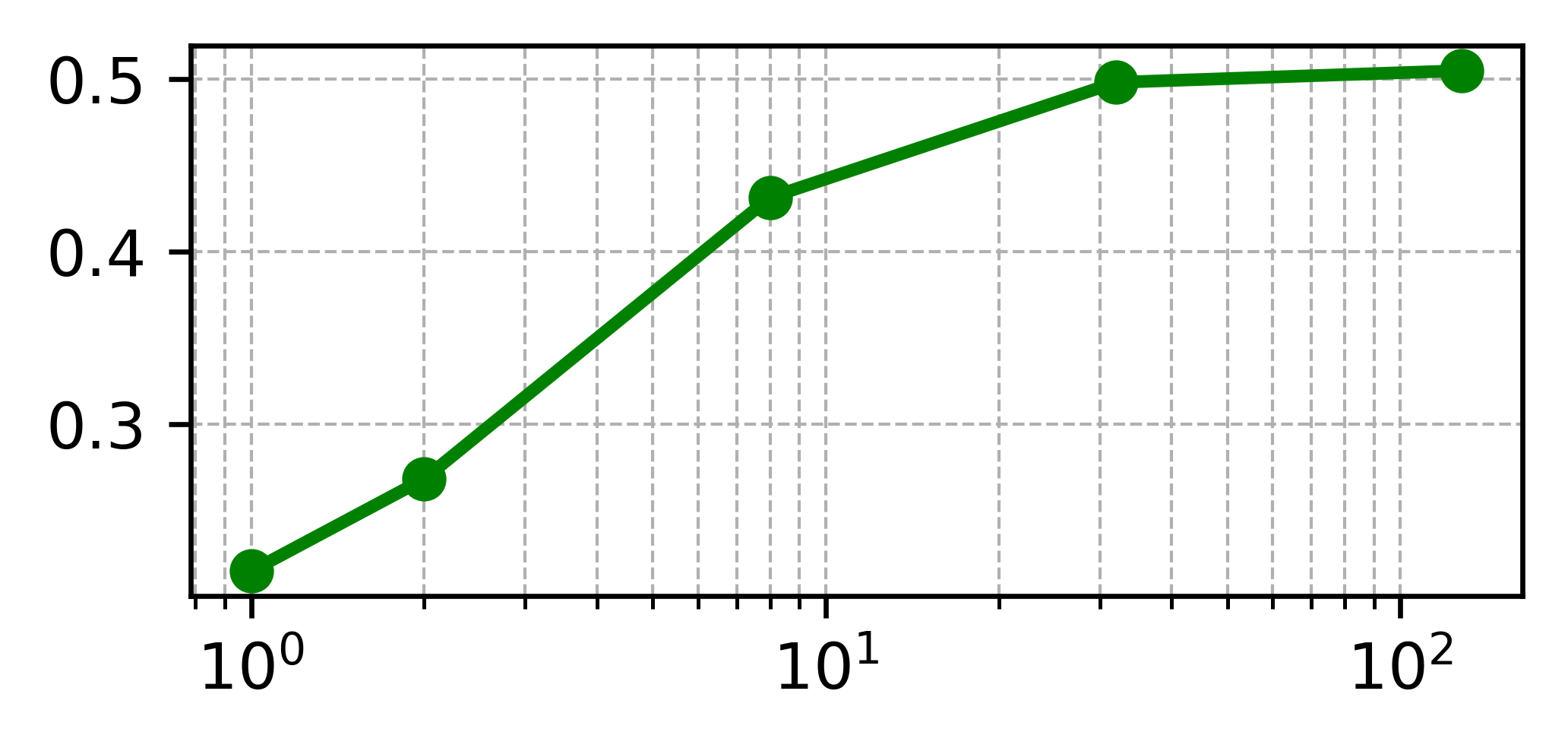}
         \put (22,-4){Canonical Angles}
        \end{overpic}
        \vspace{1mm}
        \caption{Number of canonical angles influence over Overall metric}
        \label{fig:ablation-dims}
    \end{subfigure}
    \hfill
    \begin{subfigure}[b]{0.65\columnwidth}
        \centering
        \includegraphics[width=\textwidth]{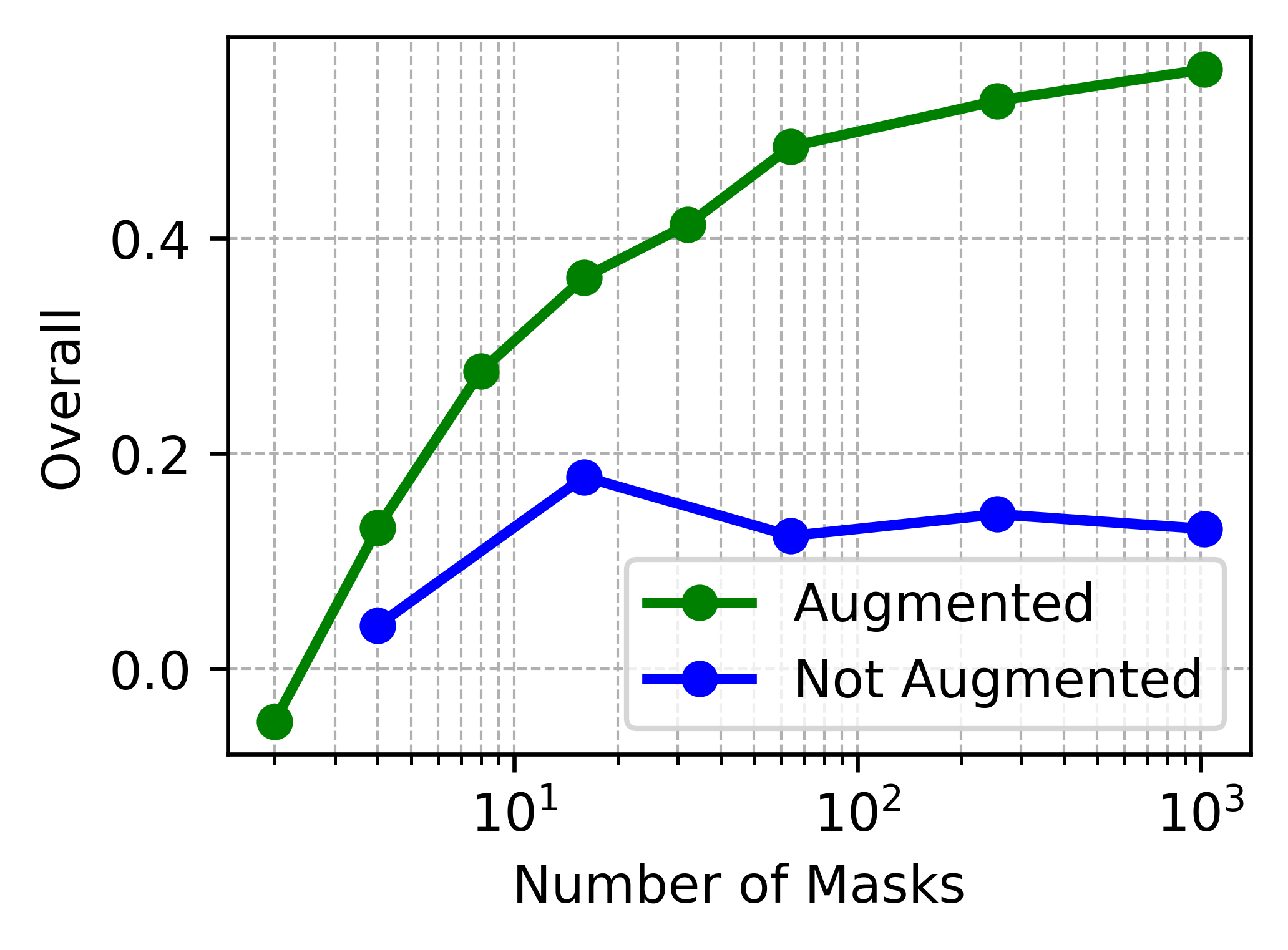}
        \caption{Number of masks and augmentations influence over overall metric}
        \label{fig:ablation-masks}
    \end{subfigure}

    
    \caption{Dependency of Overall metric with \methodabbr{} hyperparameters. The horizontal axes are set to log scale for visibility.}
    \label{fig:all_images}
    \vspace{-2mm}
\end{figure}

We examined how augmentations impact our method by switching from TrivialAugment \cite{muller_trivialaugment_2021} to RandAugment \cite{cubuk_randaugment_2020}. RandAugment allows for adjustable augmentation strength, even though the specific augmentations are random. Using 32 augmentations and 256 masks per image, we found, as seen in \cref{fig:ablation-aug-strength}, that our method remains stable up to a certain augmentation strength, beyond which it breaks down. This suggests the model can handle various augmentations as long as the image does not become unrecognizable.

Moreover, \cref{fig:ablation-masks} demonstrates our technique possesses a better convergence rate with respect to the number of masks, with 256 masks already reaching good performance. This can be traced down to the efficient masking mechanism introduced at \cref{subsubsec:anchors}. In fact, the gradient is considered the simplest version of a gradient-based interpreter~\cite{simonyan_deep_2013}, and so, all we are doing is using a quick interpreter to derive a initial probability distribution for another interpreter. Thinking from the viewpoint of chaining interpreters, we can likely consider changing the gradient for other simple options, like Grad-CAM~\cite{selvaraju_grad-cam_2016} for CNNs or Attention Rollout~\cite{abnar_quantifying_2020} for ViT~\cite{dosovitskiy_image_2021}. Also, \cref{fig:ablation-masks} shows that applying \methodabbr{} without augmentations reduces its performance significantly, showing the convergence rate is correlated with the augmentations.

Finally, we measure how many canonical angles should be used to measure the similarity between the original and occlusion subspaces. By this, we understand the impact of subspace representations to solve this problem. According to \cref{fig:ablation-dims}, there is a clear dependency with $n_c$, but also not many angles are required to reach good performance. In fact, we can see the the curve starts to saturate after 32 angles (out of 786), which already provide over $2\times$ improvement over using only 1 angle. We argue it is a strong favorable indicator for using subspace representations.

%% file: sections/conclusions.tex
In this study, we proposed a model- and class-agnostic approach for interpreting machine learning model behavior based on general augmentations and occlusions, providing robust explanations for the decision-making process of computer vision models. Our contribution lies in the application of augmentations to occluded inputs and the use of subspace representation on deep feature vectors to gauge occlusion impact with improved precision. Moreover, we enhanced the computational efficiency by transitioning the occlusion selection process from random to gradient-based. Experimental results affirm our approach's superiority over traditional methods both quantitatively and qualitatively, providing sensible explanations that effectively demystify model decisions. This work heralds significant advancements in interpretability and trustworthiness of AI systems.

%% file: supplementary/supplementary.tex
\subsection{Theory}

In this section, we present details of our method with increased mathematical formality. Some of the actual discussion and contributions may be repeated to complement the main idea.

\subsubsection{Actual causality}

\input{supplementary/causality}

\subsubsection{Occlusion Sensitivity Analysis}
\input{supplementary/occlusion}

\subsubsection{Occlusion Sensitivity Analysis with Deep Feature Vectors}
\input{supplementary/osa-das}

\subsection{Experiments}
\input{supplementary/experiments}

%% file: supplementary/causality.tex
\textit{Actual causality} \cite{halpern_causes_2005} is a framework to formally explain the way model predictions depend on input variables, what are the output causes, and how certain changes in the inputs can change the predictions. It extends counterfactual reasoning with contingencies, which means that if a Boolean function $\phi(\indices[,]{x}{n})$ changes when a variable $x_i$ is altered, then $\phi$ depends on $x_i$.

Moreover, the Degree of Responsibility $r$ is a quantification of causality~\cite{chockler_responsibility_2004}, which is based on the size $k$ of the smallest contingency required to create a counterfactual dependency~\cite{chockler_explanations_2021}, \ie, the minimal change to alter the function output.

\begin{definition}[Singleton cause] \label{def:singleton-cause}
    Let $f$ be a machine learning model and $x$ an input, an entry $x_{\indices{i}{n}}$ (think a pixel) is a cause of $f(x)$ \textit{if and only if} there is a subset $\chi \subset x$ such that the following hold~\cite[p.~3]{chockler_explanations_2021}:
    \begin{enumerate}
        \item $x_{\indices{i}{n}} \not\in \chi$
        \item output invariance to masking of $\chi$: $$\text{ Let } \chi' \subset \chi, m \in \mathbb{R}, \text{ then } \chi' = m \implies \Delta f = 0$$
        \item output dependency to masking of $x_{\indices{i}{n}}$: $$\text{ Let } m \in \mathbb{R}, \text{ then } \chi = x_{\indices{i}{n}} = m \implies \Delta f \neq 0$$
    \end{enumerate}
\end{definition}

\begin{definition}[Cause witness] \label{def:cause-witness}
    If a subset $\chi \in x$ and entry $x_{\indices{i}{n}}$ satisfy \cref{def:singleton-cause}, then we say $\chi$ is a witness to the fact $x_{\indices{i}{n}}$ is a cause of $x$~\cite[p.~3]{chockler_explanations_2021}.
\end{definition}

\begin{definition}[Simplified Degree of Responsibility] \label{def:simp-resp}
    If a subset $\chi \in x$ and entry $x_{\indices{i}{n}}$ satisfy the definition of Singleton Cause from~\cite{chockler_explanations_2021}, then 
    \begin{equation}
        r(x_{\indices{i}{n}}|x,f) = \frac{1}{1 + k},
    \end{equation}
    where $k$ is the size of the minimal witness, which refers to the smallest subset of input variables that, when changed, can demonstrate that a particular input variable has an effect on the output of a function.
\end{definition}

In our terminology, interpreters are algorithms which process a model and a single input, and output an explanation. In computer vision, a valid explanation could be an attribution heatmap over the original input image. Next, we formally define this concept.

\begin{definition}[Explanation] \label{def:explanation}
    Let $f$ be a machine learning model and $\mathbf{x}$ an input with output $f(\mathbf{x})$, and $\mathbf{S} = \mathbf{x} \odot \mathbf{M}$ a masked subset of the input. Then, the explanation $\mathbf{E}$ of the model $f$ given input $\mathbf{x}$ is the minimal subset which maintains the output~\cite{chockler_explanations_2021} 
    \begin{equation}
        \mathbf{E}(f|\mathbf{x}) = \min_{|\mathbf{S}|} \mathbf{S} : f(\mathbf{S}) = f(\mathbf{x}), |\mathbf{S}| > 0
    \end{equation}

    where $|\cdot|$ is the number of items in the set, \eg, the number of non-masked pixels
\end{definition}

\begin{remark}[Triviality]
    The explanation must not be a null tensor.
    Any model will output a prediction for the null tensor, however this would be a trivial explanation for all inputs sharing the same prediction. For example, if a model outputs “dog” for the null tensor, then all images of dogs would have an empty heatmap as explanation.
\end{remark}

\begin{remark}[Non-uniqueness]
    The explanation may not be unique. 
    Inputs might have symmetries or repetitions, leading to multiple viable subsets of the same size.
\end{remark}

However, computing an explanation is NP-complete and real interpreters will output approximate explanations~\cite{chockler_explanations_2021}.

\begin{definition}[Approximate Explanation] \label{def:approx-expl}
    Let $f$ be a machine learning model and $\mathbf{x}$ an input with output $f(\mathbf{x})$. The approximate explanation $\Tilde{\mathbf{E}}$ of model $f$ given input $\mathbf{x}$ is the probability distribution indicating if the input entry belongs to the explanation $\mathbf{x}$.
    \begin{equation}
        \Tilde{\mathbf{E}}_{\indices{i}{n}} = p(\mathbf{x}_{\indices{i}{n}} \in \mathbf{x})
    \end{equation}
\end{definition}

\begin{remark}[Normalization] \label{remark:approx-expl-norm}
    \begin{equation}
        \sum_{\indices{i}{n}} \Tilde{\mathbf{E}}_{\indices{i}{n}} = 1
    \end{equation}
\end{remark}

\begin{remark}[Approximate Explanation and Degree of responsibility] \label{remark:approx-expl-and-resp}
    While explanations can be seen as a binary inclusion mask, \ie, the mask is 1 if the entry belongs to the explanation. However, the degree of responsibility is a measure between 0 and 1, so it can be seen as a proxy for probability.
\end{remark}

\begin{remark}[Composition] \label{remark:expl-composition}
    Given the probabilistic nature of an approximate explanation, the composed approximate explanation can be built when multiple approximate explanations are available (\eg, when multiple interpreters can be used). The composed explanation can be built by simple summation and renormalization following \cref{remark:approx-expl-norm}.
\end{remark}

\begin{definition}[Minimal Size] \label{def:minimal-size}
    Let $\Tilde{\mathbf{E}}$ be the approximate explanation of model $f$ given input $\mathbf{x}$. The explanation minimal size is defined as

    \begin{equation}
        s_{min}(\Tilde{\mathbf{E}}(f|\mathbf{x})) = \frac{|\mathbf{S}|}{|\mathbf{x}|}, \text{ with } f(\mathbf{S}) \approx f(\mathbf{x})
    \end{equation}
\end{definition}

\begin{remark}[Minimal Size metric] \label{remark:minimal-size-metric}
    \cref{def:minimal-size} is a viable explanation metric. It can be measured by \cref{alg:minimal-size}. In that sense, an explanation with low minimal size indicates the most substantial region for the model was reached. 
    
\end{remark}

\begin{algorithm}
\caption{Minimal Size Metric Computation} \label{alg:minimal-size}
\begin{algorithmic}
\Require $\mathbf{x} \gets$ image, $f \gets$ model, $\mathbf{H} \gets$ heatmap, $s \gets$ number of steps, $\delta \gets$ tolerance

\For{$i \gets 1$ to $|\mathbf{x}|$ in $s$ steps}
\State $\mathbf{S} \gets$ top $i$ pixels from $\mathbf{x}$ based on $\mathbf{H}$
\If{$||f(\mathbf{x}) - f(\mathbf{S})||_1 \leq \delta$}
    \Return $|\mathbf{S}| / |\mathbf{x}|$
\EndIf
\EndFor
\end{algorithmic}
\end{algorithm}



%% file: supplementary/occlusion.tex
Occlusion computes explanation heatmaps by replacing image regions with a given baseline (masking it to 0), and measuring the score difference in the output~\cite{fleet_visualizing_2014, petsiuk_rise_2018}.

\begin{proposition}[Occlusion degree of responsibility] \label{prop:occlusion-resp}
    Let $f$ be a model which outputs a probability score $p \in [0, 1]$, $\mathbf{x}$ an input and $\mathbf{M}$ as binary mask with the same shape as $\mathbf{x}$, and $\odot$ be the Hadamard product. Then, the degree of responsibility of the masked region is $$r = 1 - \frac{p(\mathbf{x} \odot \mathbf{M})}{p(\mathbf{x})}$$

\begin{proof}
    Let the degree of responsibility be $r = \frac{1}{1 + k}$  (\cref{def:simp-resp}), the factor $k$ represents the size of the minimal witness~\cite{chockler_explanations_2021}, which should be $0$ for relevant causes and $\infty$ for irrelevant ones, \ie, $k \in [0, +\infty)$.

    First, assume that for every image, there is a defined region $\mathbf{S}$ where it's minimal witness has size $0$, \ie, when we mask all of the image keeping nothing but $\mathbf{S}$, the prediction score $p$ output by the model is unaltered. Moreover, assume that the opposite action is also true: masking $\mathbf{S}$ while keeping any other part of the image will lead to a prediction collapse. Simply put, $$p(\mathbf{x} - \mathbf{S}) = 0 \iff p(\mathbf{S}) = p(\mathbf{x})$$

    Conversely, if $\mathbf{S}'$ is an irrelevant area, masking it should render no change, $$p(\mathbf{x} - \mathbf{S}') = p(\mathbf{x}) \iff p(\mathbf{S}') = 0$$

    From such assumption, we should expect that the masked region minimal witness~\cite{chockler_explanations_2021} size must be proportional to the score $p(\mathbf{S})$ of keeping only the cause $\mathbf{S}$, and inversely proportional to the score $p(\mathbf{S}')$ of keeping only an unimportant region $\mathbf{S}'$.

    Thus, we can say $$k \propto \frac{p(\mathbf{S})}{p(\mathbf{S}')} \equiv \frac{p(\mathbf{x} - \mathbf{S})}{|p(\mathbf{x}) - p(\mathbf{x} - \mathbf{S}')|} \equiv \frac{p(\mathbf{x} \odot \mathbf{M})}{|p - p(\mathbf{x} \odot \mathbf{M})|}, $$
    which should be $0$ when the cause is masked and diverge when masking an unimportant region (score does not change).

    Finally, we can effectively ignore the modulo assuming $p(\mathbf{x}) \ge p(\mathbf{x} \odot \mathbf{M})$, and 
    \begin{equation}
        \begin{split}
            r &= \frac{1}{1 + k} \\
              &= \frac{1}{1 + \frac{p(\mathbf{x} \odot \mathbf{M})}{p - p(\mathbf{x} \odot \mathbf{M})}} \\
              &= \frac{p(\mathbf{x}) - p(\mathbf{x} \odot \mathbf{M})}{p(\mathbf{x})} \\
              &= 1 - \frac{p(\mathbf{x} \odot \mathbf{M})}{p(\mathbf{x})} \\
        \end{split}
    \end{equation}
\end{proof}
\end{proposition}

\begin{proposition}[Occlusion approximate explanation] \label{prop:occlusion-approx-expl}
    The approximate explanation of an input for a single mask is $$\Tilde{\mathbf{E}}^\mathbf{M}(f|\mathbf{x}) = (1 - \mathbf{M}) (1 - \frac{p(\mathbf{x} \odot \mathbf{M})}{p(\mathbf{x})})$$

    \begin{proof}
        Given $\mathbf{M}$ is a binary mask, $1 - \mathbf{M}$ is the inverse mask, \ie, the masked region is set to $1$.
        Then, from \cref{prop:occlusion-resp} and \cref{remark:approx-expl-and-resp} we say the probability of the cause belonging to the masked region is equal the term $1 - \frac{p(\mathbf{x} \odot \mathbf{M})}{p(\mathbf{x})}$.
    \end{proof}
\end{proposition}

\begin{lemma}[Occlusion Sensitivity Analysis (OSA)] \label{lemma:osa}
    The non-normalized general occlusion sensitivity analysis is the combination of individual occlusion explanations.
    \begin{equation} \label{eq:occlusion}
    \Tilde{\mathbf{E}}(f|\mathbf{x}) = \sum_{i} (1 - \mathbf{M}_i) (1 - \frac{p(\mathbf{x} \odot \mathbf{M}_i)}{p(\mathbf{x})})
    \end{equation}

    \begin{proof}
        Direct from \cref{prop:occlusion-approx-expl} and \cref{remark:expl-composition}
    \end{proof}
\end{lemma}

OSA is a simple example of a perturbation-based method, in which the explanation is a composition of output scores relative variations for each masked input. \cref{prop:occlusion-resp} defines a way of computing the responsibility of a singular mask, \ie, the probability it belongs to the cause, and \cref{alg:occlusion} shows how to compose it into an approximate explanation. OSA pseudocode can be found in the supplementary material.

\begin{algorithm}
\caption{Occlusion Sensitivity Analysis} \label{alg:occlusion}
\begin{algorithmic}
\Require $\mathbf{x} \gets \text{image}, f \gets \text{model}$
\State $n \gets \text{number of masks}$
\State $l \gets \text{mask size}$
\State $p \gets f(\mathbf{x})$
\State $\mathbf{H} \gets 0$
\For{$i \gets 1$ to $n$}
    \State $\mathbf{M} \gets \text{mask}(i, \mathbf{x}.shape, l)$ \Comment{any mask generator}
    \State $\mathbf{x}^\mathbf{M} \gets \mathbf{x} \odot \mathbf{M}$
    \State $p^\mathbf{M} \gets f(\mathbf{x}^\mathbf{M})$
    \State $\mathbf{H} \gets \mathbf{H} + (1 - \mathbf{M}) (1 - \frac{p^\mathbf{M}}{p})$ \Comment{compose (\cref{remark:expl-composition})}
\EndFor
\end{algorithmic}
\Return $\frac{\mathbf{H}}{\sum \mathbf{H}}$ \Comment{normalize explanation}
\end{algorithm}

However, notice the formulation at \cref{lemma:osa} is different to the more traditional one defined by~\cite{petsiuk_rise_2018} in equation 6. The difference is mostly due to the different assumptions we took, but they can be shown to be proportionally equivalent, differing only by a constant. However, the formulation at \cref{lemma:osa} will be important for the methods we will propose next.

%% file: supplementary/osa-das.tex
\cref{lemma:osa} is a class-specific algorithm. However, most machine learning models actually output general vectors, also known as \modeloutput{}, which encode the input through the model. These vectors are then later processed to obtain the probability score of a single feature (class).

\begin{proposition}[Representation degree of responsibility] \label{prop:repr-resp}
    Let $f$ be a model which outputs a vector $f(\mathbf{x}) = \mathbf{v} = (v_i), i \in [m]$. The degree of responsibility of the masked region is 
    $$r = \frac{||\mathbf{v} - \mathbf{v}(\mathbf{x} \odot \mathbf{M})||_p}{||\mathbf{v}||_p}, $$ 

    where $||\mathbf{v}||_p$ stands for the ${\ell_{p}}$-norm

    \begin{proof}
        Let $f$ be a model which outputs a scalar probability score $p(\mathbf{x}) \in [0,1]$. Then, from \cref{prop:occlusion-resp}, the degree of responsibility is 
        \begin{equation}
            \begin{split}
                r &= 1 - \frac{p(\mathbf{x} \odot \mathbf{M})}{p(\mathbf{x})} \\
                  &= \frac{p(\mathbf{x}) - p(\mathbf{x} \odot \mathbf{M})}{p(\mathbf{x})} \\ 
                  &= \frac{f(\mathbf{x}) - f(\mathbf{x} \odot \mathbf{M})}{f(\mathbf{x})} \\
            \end{split}
        \end{equation}

        Now, notice that \cref{prop:occlusion-resp} assumes $f(\mathbf{x} \odot \mathbf{M}) \le f(\mathbf{x})$ and $p(\mathbf{x}) \ge 0$. Then, we can extend this concept to a new $f$ which outputs vectors by 
        \begin{equation}
            \begin{split}
                \frac{f(\mathbf{x}) - f(\mathbf{x} \odot \mathbf{M})}{f(\mathbf{x})} &= \\
                \frac{|f(\mathbf{x}) - f(\mathbf{x} \odot \mathbf{M})|}{|f(\mathbf{x})|} &= \\ 
                \frac{||\mathbf{v} - \mathbf{v}(\mathbf{x} \odot \mathbf{M})||_p}{||\mathbf{v}||_p} \\
            \end{split}
        \end{equation}
    \end{proof}
\end{proposition}

\begin{remark}[Occlusion sensitivity analysis as a special case]
    The vector extension in \cref{prop:repr-resp} also shows that \cref{lemma:osa} is a special case when we wish to analyze one particular feature of $f(\mathbf{x})$, so this can be thought as a generalization of said method.
\end{remark}

\begin{lemma}[Representation Occlusion Sensitivity Analysis] \label{lemma:rosa}
    The natural extension to dealing with representations reintroduces \cref{lemma:osa} with the only change in the degree of responsibility calculation. 

    \begin{equation}
        \Tilde{\mathbf{E}}(f|\mathbf{x}) = \sum_{i} (1 - \mathbf{M}_i) \frac{||\mathbf{v} - \mathbf{v}(\mathbf{x} \odot \mathbf{M})||_p}{||\mathbf{v}||_p}    
    \end{equation}

    \begin{proof}
        Analogous to \cref{lemma:osa}.
    \end{proof}
    
\end{lemma}

\begin{remark}[Representation Occlusion Sensitivity Analysis]
    The natural extension to dealing with representations from \cref{lemma:rosa} reintroduces \cref{lemma:osa} with the only change in the degree of responsibility calculation.
\end{remark}

\subsubsection{Occlusion Sensitivity Analysis with Deep Feature Augmentation Subspaces} \label{subsubsec:rosa-sm}


While \cref{lemma:rosa} outlines a general method to determine the degree of responsibility, its sole dependence on occlusion may not capture the nuanced relationships inherent in deep learning models. Recognizing the vital role of data augmentation in training and viewing occlusion as a form of augmentation, we propose a shift from simple vector comparisons to a detailed analysis between two subspaces. A subspace here denotes a segment of the \modeloutput{} space defined by an occluded image and its augmentations. We strive to assess the similarity between each ``occlusion subspace'' and the ``reference subspace'', which is formed by the original image and its augmentations. Extending \cref{lemma:rosa}, we compare the size difference between two subspaces, focusing on their orthogonal degree, and measure the canonical angles between subspaces derived from varied transformations on the original and occluded images.

\begin{proposition}[Subspace degree of responsibility] \label{prop:subspace-resp}
    The degree of responsibility between subspaces is the orthogonal degree between them, \ie, 

    \begin{equation}
        r({\mathbf{M}}) = 1 - \sum_i^{n_c} (\sigma_{i}(\mathbf{V}^T \mathbf{V}_\mathbf{M}))^2
    \end{equation}
    
\begin{proof}
    We extend the idea (and the notation) of difference of vectors to difference of subspaces

    \begin{equation}
    \begin{split}
        r &= \frac{||\mathbf{v} - \mathbf{v}(\mathbf{x} \odot \mathbf{M})||_2}{||\mathbf{v}||_2} \\
        &\equiv \frac{|\mathcal{V} - \mathcal{V}_\mathbf{M}|}{|\mathcal{V}|} \\
        &= \frac{|\mathcal{V} - \mathcal{V}_\mathbf{M}|}{|\mathcal{V} - \mathbf{0}|} \\
        &= \frac{|\mathcal{V} - \mathcal{V}_\mathbf{M}|}{1} \\
        &= |\mathcal{V} - \mathcal{V}_\mathbf{M}| \\
        &= 1 - simi(\mathcal{V},\mathcal{V}_\mathbf{M}) \\
        &= 1 - \sum_i^{n_c} (\sigma_{i}(\mathbf{V}^T \mathbf{V}_\mathbf{M}))^2 \\
    \end{split} 
    \end{equation}

    where $|\mathcal{A} - \mathcal{B}|$ is a subspace distance, \ie, the orthogonal degree between $\mathcal{A}$ and $\mathcal{B}$~\cite{fukui_difference_2015,fukui_subspace_2020}. 
    $\mathbf{V}$ and $\mathbf{V}_\mathbf{M} \in {\mathbb{R}}^{k{\times}d}$ are the orthonormal basis of the subspaces $\mathcal{V}$ and $\mathcal{V}_\mathbf{M}$ respectively.
\end{proof}

\begin{theorem}[\methodfullname{}] \label{lemma:osa-das}
    The natural extension to dealing with deep feature augmentation subspaces reintroduces \cref{lemma:rosa} with the only change in the degree of responsibility calculation. 

    \begin{equation}
        \Tilde{\mathbf{E}}(f|\mathbf{x}) = \sum^{n_m}_{i} (1 - \mathbf{M}_i) (1 - \sum_{j}^{n_c} \sigma^2_{j}(\mathbf{V}^T \mathbf{V}_{\mathbf{M}_{i}}))
    \end{equation}

    \begin{proof}
        Analogous to \cref{lemma:rosa}.
    \end{proof}
    
\end{theorem}

\end{proposition}

%% file: supplementary/experiments.tex
    \begin{table*}[!htbp]
    \centering
    \caption{Metric scores on ImageNet for ResNet-50, ViT-B and Swin-V2. For deletion and minimal size, lower is better ($\downarrow$). For insertion, higher is better ($\uparrow$). \textbf{Bold} represents the best metric for a given model, while \underline{underline} is the second best.}
    \begin{tabular}{llcccc}
    \toprule
    Method & \textbf{Model} & \textbf{Minimal Size ($\downarrow$)} & \textbf{Deletion ($\downarrow$)} & \textbf{Insertion ($\uparrow$)} \\
    \midrule
    \multirow{3}{*}{Grad-CAM~\cite{selvaraju_grad-cam_2016}} & ResNet-50 & 0.532 & \underline{0.181} & 0.174 \\
    & ViT-B     & 0.512 & 0.340 & 0.415 \\
    & Swin-V2   & 0.502 & \underline{0.374} & 0.279 \\
    \midrule
    \multirow{3}{*}{Integrated Gradients~\cite{sundararajan_axiomatic_2017}} & ResNet-50 & 0.522 & \textbf{0.086} & 0.248  \\
     & ViT-B     & 0.541 & \textbf{0.223} & 0.378  \\
     & Swin-V2   & 0.492 & \textbf{0.177} & 0.393 \\
    \midrule
    \multirow{3}{*}{Occlusion~\cite{fleet_visualizing_2014}} & ResNet-50 & \underline{0.255} & 0.278 & \underline{0.456} \\
    & ViT-B     & \textbf{0.343} & \underline{0.295} & \textbf{0.521} \\
    & Swin-V2   & \textbf{0.155} &  0.410 & \textbf{0.670} \\
    \midrule
    \multirow{3}{*}{Ours}      & ResNet-50 & \textbf{0.137} & 0.291 & \textbf{0.530} \\
    & ViT-B     & \underline{0.346} & 0.311 & \underline{0.507} \\
    & Swin-V2   & \underline{0.210} & 0.387 & \underline{0.579} \\
    \bottomrule
    \end{tabular}
    \label{tab:metric-comparisons-full}
\end{table*}

All models used Imagenet-1k weights provided by torchvision. Grad-CAM~\cite{selvaraju_grad-cam_2016} uses Captum~\cite{kokhlikyan_captum_2020} GuidedGradCam implementation. Grad-CAM is set to track the last convolutional layer on ResNet-50~\cite{he_deep_2016} and the least BatchNormalization layer on ViT-B~\cite{dosovitskiy_image_2021} and Swin-V2~\cite{liu_swin_2022}. Integrated Gradients~\cite{sundararajan_axiomatic_2017} uses Captum~\cite{kokhlikyan_captum_2020} IntegratedGradients implementation. We use a null baseline, computing the integral on 128 steps with Gauss–Legendre quadrature. Occlusion Sensitivity Analysis~\cite{fleet_visualizing_2014,uchiyama_visually_2023} uses Captum~\cite{kokhlikyan_captum_2020} Occlusion, which is implemented with a sliding window of binary masks with $32 \time 32$ pixels and stride of $1$. This is equivalent to approximately $9216$ masks per image. Quantitative results for each interpreter on each model can be found at \cref{tab:metric-comparisons-full}.